\documentclass[conference]{IEEEtran}
\usepackage{amsmath,amssymb,booktabs,array,graphicx,cite,url}

\title{Multimodal Progress-Residual Fusion for Anticipating Navigation Retreat in Robotic Flexible Endoscopy}
\author{\IEEEauthorblockN{Chi Kit Ng\textsuperscript{*}, Tianchun Wu\textsuperscript{*}, Jiachen Sun\textsuperscript{*},
Yuming Sun, Jianru Zhang, Hongliang Ren\textsuperscript{$\dagger$}}
\IEEEauthorblockA{\footnotesize\textsuperscript{*}Equal contribution. \textsuperscript{$\dagger$}Corresponding author.}}
\begin{document}
\maketitle

\begin{abstract}
Flexible robotic endoscopes navigate tortuous anatomy where friction, deformation, and wall interaction separate actual distal movement from the response expected under an insertion command. Operators interpret this discrepancy together with the endoscopic scene and retreat for reorientation. Learning this judgment is a robotics problem of inferring navigation intent from multimodal histories. We introduce Multimodal Progress-Residual Fusion (MPRF), a causal model that learns their temporal relationship to emulate the operator's decision to retreat. A nominal-progress model predicts expected motion; a temporal predictor fuses its electromagnetic (EM) residual with commands, EM pose, and video. In 150 colon-phantom episodes, MPRF achieves $0.580\pm0.007$ AUPRC across three seeds, compared with 0.511 for the strongest reduced-modality control. Complete multimodal input supplies the principal ranking advantage; the residual adds an explicit expected--actual response representation with a small, uncertain increment over raw fusion. Alarm analysis relates warning timing to the recorded retreat decision. These results provide a basis for learning human-like navigation-intent judgment, with prospective navigation success and distal-force exposure as the intended measures of robotic utility.
\end{abstract}

\section{Introduction}
Flexible and continuum endoscopes provide access through natural orifices and curved anatomical passages while limiting the access incision~\cite{burgner2015continuum,runciman2019soft}. Robotic actuation adds distal dexterity, stabilizes viewpoints, and supports repeatable navigation~\cite{zhang2020novel,ma2019autonomous}. During teleoperation, these capabilities still depend on an operator who interprets whether intended motion produces useful progress. The operator repeatedly chooses between continued insertion, orientation adjustment, and withdrawal. Recognizing an approaching change of strategy is therefore an important component of assistance alongside motion generation.

\begin{figure*}[t]
\centering\includegraphics[width=\textwidth]{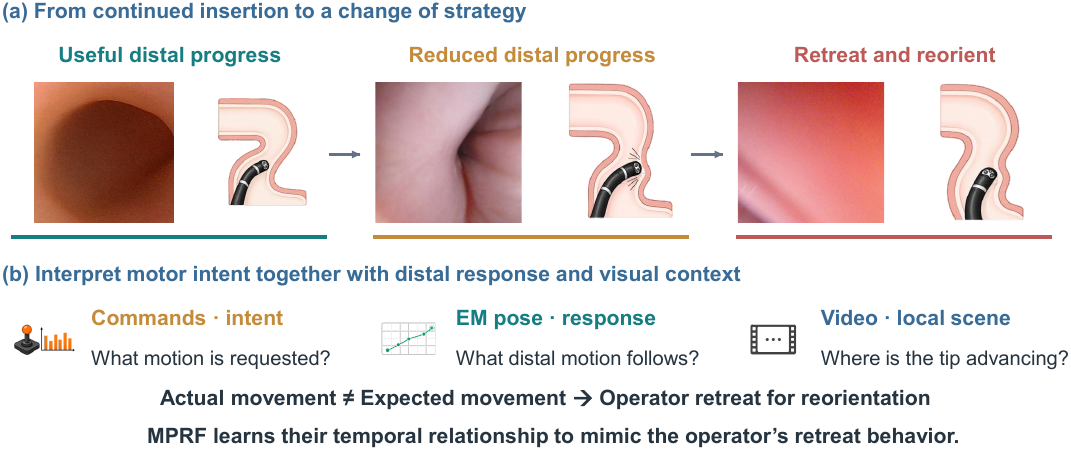}
\caption{Robotics motivation and sensing roles. (a) A discrepancy between actual and expected movement motivates operator retreat for reorientation. (b) Commands encode intent, EM pose measures distal response, and video supplies local context. MPRF learns their temporal relationship to emulate the operator's retreat decision.}
\label{fig:motivation}
\end{figure*}

The robotics motivation is the discrepancy between \emph{actual movement} and \emph{expected movement}, followed by an operator's decision to retreat for reorientation (Fig.~\ref{fig:motivation}). Friction, backlash, deformation, and environmental interaction separate proximal actuation from distal response~\cite{webster2010design,poignonec2020towards}. Sustained insertion produces forward translation, local bending, or little progress; limited displacement also accompanies deliberate inspection. The judgment to persist or retreat therefore depends on the temporal relationship between commanded motion, measured response, and visual context. Learning this relationship gives the robot a representation of the operator's navigation intent.

MPRF targets this component of human judgment through retreat anticipation. Its advisory probability supports inspection and reorientation, with action authority retained by the operator, consistent with shared control in medical robotics~\cite{yang2017medical,dupont2021decade}. Recorded retreat decisions provide an initial learning signal. The longer-term objective is improved navigation through multimodal intent inference; successful navigation and physical interaction provide the corresponding task-level endpoints.

Existing approaches improve contact observability, estimate continuum state, or compare expected and measured body response~\cite{ha2022contact,wang2024sensing}. Temporal video--kinematic models also predict surgical activities~\cite{weerasinghe2024multimodal}. The research gap addressed here is a short-horizon behavioral transition during flexible-endoscope navigation: relating \emph{what the operator commands}, \emph{how the tip responds}, and \emph{what the camera observes} to the decision to retreat. An isolated stream admits several plausible interpretations. Their temporal relationship supplies the context needed to distinguish sustained progress from an attempt approaching abandonment.

The recorded transition from manual navigation to the retreat controller provides weak behavioral supervision. This operational definition preserves the operator's actual decision and gives a precise log index for causal training. It encompasses recovery and repositioning, with timing that reflects operator judgment. We consequently evaluate both the trained horizon and an expanded matching window for the same predictions.

MPRF makes the intent--response relationship explicit. A first stage learns nominal EM and visual increments from normal and successful navigation. The difference between expected and observed EM change becomes a progress residual. A causal recurrent model fuses this residual with the original commands, pose, and visual features so that context determines its predictive meaning.

The study contributes a log-derived formulation of retreat anticipation, a two-stage residual representation conditioned on multimodal context, and a matched evaluation that separates sensor complementarity, residual contribution, and alarm timing. Episode-level resampling quantifies the sensitivity of the principal ranking contrasts; probability traces explain how the fixed alarm rule converts scores into warnings.

\section{Related Work}
\subsection{Robotic Endoscopy and Operator Assistance}
Continuum mechanisms accommodate curved anatomical paths through distributed deformation~\cite{webster2010design,burgner2015continuum}. Robotic endoscopes combine flexible actuation with interfaces for distal steering and insertion~\cite{zhang2020novel}. Sensor-guided colonoscopy and autonomous endoscopic control address navigation and task execution~\cite{chen2009sensor,ma2019autonomous}. These advances motivate assistance that also recognizes the state of an ongoing human-controlled attempt. Within the autonomy spectrum~\cite{yang2017medical,dupont2021decade}, MPRF predicts a forthcoming operator transition and supplies an advisory probability. Its supervision derives directly from recorded control-state changes.

\subsection{Sensing and Expected Response}
Data-driven contact localization improves observability of continuum interaction~\cite{ha2022contact}. Endoscopic vision also reveals transmission effects, including backlash~\cite{poignonec2020towards}. Wang \emph{et al.} compared expected and observed soft-body response for proprioception and contact detection~\cite{wang2024sensing}. This principle motivates our progress residual, with a different target: future operator retreat. Raw multimodal context remains essential because friction, steering, tracker variation, and successful exploration all influence an observed discrepancy. MPRF learns its behavioral interpretation from causal histories.

\subsection{Multimodal Temporal Prediction}
Surgical data science integrates heterogeneous procedural signals to support intervention~\cite{maier2017surgical}. EndoNet learns visual representations for phase and tool recognition~\cite{twinanda2016endonet}; SV-RCNet incorporates temporal dependencies for workflow recognition~\cite{jin2017sv}. Multimodal transformers combine video and robot kinematics for activity prediction~\cite{weerasinghe2024multimodal}. Here the event is an imminent navigation-strategy change. Commands provide the intent needed to interpret EM motion, while video contextualizes the distal scene. Matched modality controls test the value of retaining each stream, and a complete raw-fusion control isolates the additional residual channel.

\section{Method}
\subsection{Causal Inputs and Weak Behavioral Labels}
The synchronized record contains image $I_t$, EM pose $E_t\in\mathbb{R}^{7}$ (position and quaternion components), and four logged motor commands $A_t$. The two insertion channels have equal recorded values; two further channels command bending. Training-only statistics standardize commands and EM pose. A frozen ImageNet-pretrained MobileNetV3-Small~\cite{howard2019searching} encodes each image into 576 features, followed by a training-fitted 32-dimensional PCA projection $V_t$. Missing images use causal carry-forward, initialized with a zero backbone vector before PCA.

An eligible attempt is a manual-navigation segment of at least 32 rows, consistent with the shared extraction protocol. If retreat starts at row $b$, pre-onset endpoints receive
\begin{equation}
y_t=\mathbb{1}[b-10\leq t<b].
\label{eq:label}
\end{equation}
The input ends before $b$; controller state supplies labels only. Other eligible attempts supply negative examples. MPRF uses 16-row histories. The protocol converts row counts at 20~Hz: 16 rows correspond to a nominal 0.8~s and the 10-row horizon to 0.5~s. Recorded timestamps have a median increment of 50~ms with occasional gaps; model horizons and alarm rates retain this explicit row-based convention.

\begin{figure*}[t]
\centering\includegraphics[width=\textwidth]{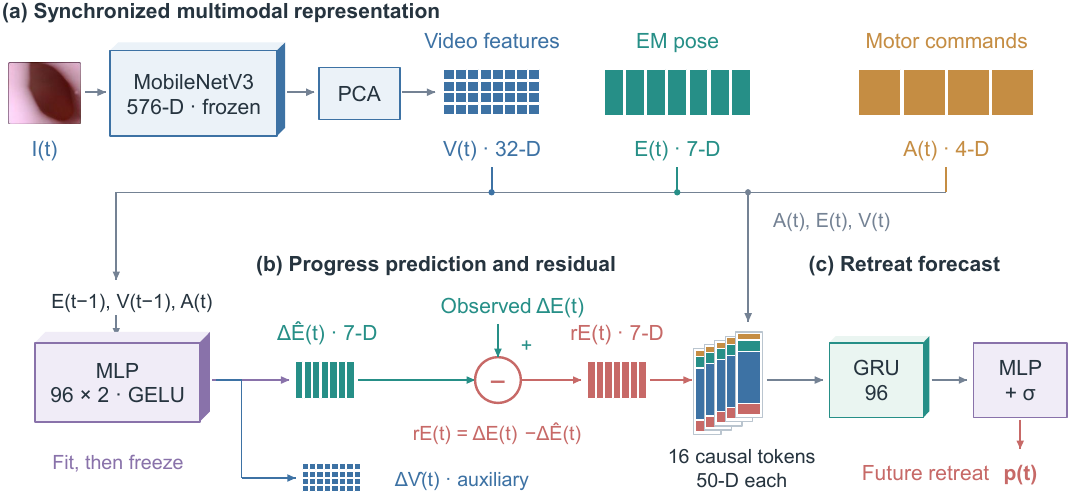}
\caption{MPRF architecture. (a) Frozen visual encoding and training-fitted PCA provide 32 visual features alongside EM pose and motor commands. (b) Stage A predicts nominal EM and visual increments from prior state and current command. Subtracting predicted EM progress from observed progress gives the residual; the visual head supplies auxiliary supervision. Stage A fits Normal and Pass attempts with dimension-averaged $L_1$ losses, then stays frozen. (c) Current A/E/V and the residual form 50-dimensional tokens. Stage B encodes 16 causal tokens with a GRU and predicts future retreat through layer normalization, a 48-unit GELU head, dropout, and a sigmoid.}
\label{fig:method}
\end{figure*}

\subsection{Nominal Progress and Residual Fusion}
Stage A learns the response expected during Normal navigation and passable narrow-lumen navigation (Fig.~\ref{fig:method}). A two-layer MLP, each hidden layer containing 96 units and GELU activation, predicts
\begin{equation}
(\widehat{\Delta E}_t,\widehat{\Delta V}_t)
=g_\phi(E_{t-1},V_{t-1},A_t),
\end{equation}
where $\Delta E_t=E_t-E_{t-1}$ and $\Delta V_t=V_t-V_{t-1}$. Its training objective is
\begin{equation}
\mathcal L_A=\frac{\|\Delta E_t-\widehat{\Delta E}_t\|_1}{7}
+0.5\frac{\|\Delta V_t-\widehat{\Delta V}_t\|_1}{32}.
\end{equation}
Each term is also averaged across the minibatch. Quaternion-component increments serve as empirical model coordinates. After nominal-progress fitting, Stage A stays frozen.

The residual and fused token are
\begin{equation}
r_t^E=\Delta E_t-\widehat{\Delta E}_t,\qquad
z_t=[A_t,E_t,V_t,r_t^E]\in\mathbb{R}^{50}.
\end{equation}
At the first row of an attempt, the residual is initialized to zero. The residual expresses deviation from learned nominal response; the retained raw modalities contextualize that deviation. The visual-increment head provides auxiliary supervision, and Stage B receives the original visual features.

A single-layer, width-96 GRU processes $z_{t-15:t}$. Its final state passes through layer normalization, a 48-unit GELU layer, dropout 0.15, and a sigmoid to produce $p_t$. Stage B minimizes positive-weighted binary cross-entropy. Validation selects the checkpoint and the threshold/persistence alarm rule; a 20-row refractory interval separates emissions.

\section{Experimental Design}
\subsection{Platform, Recording Conditions, and Partition}
A single operator navigated a robotic flexible endoscope through a colon phantom with synchronized video, distal EM tracking, and command logging. The 150 episodes comprise 50 per condition: \emph{Normal navigation} uses a wide lumen; \emph{Narrow-lumen (can pass)} uses a constricted opening larger than the endoscope diameter; \emph{Narrow-lumen (cannot pass)} uses an opening smaller than the endoscope diameter. N/P/F retain the recording labels Normal/Pass/Failure for these conditions. They describe the physical setting; retreat-controller transitions supply the predictive target.

The 30,662 aligned rows contain 112 raw retreat transitions and 96 eligible events. The episode-disjoint training, validation, and evaluation partitions contain 91/28/31 recordings, 118/37/41 eligible attempts, and 56/19/21 events. One short recording contributes zero eligible attempts. The evaluation partition supplies 5,392 causal endpoints, including 210 positives (3.89\%). Full-recording summaries characterize the acquisition conditions; predictive metrics use censored pre-retreat windows.

\begin{figure*}[t]
\centering\includegraphics[width=\textwidth]{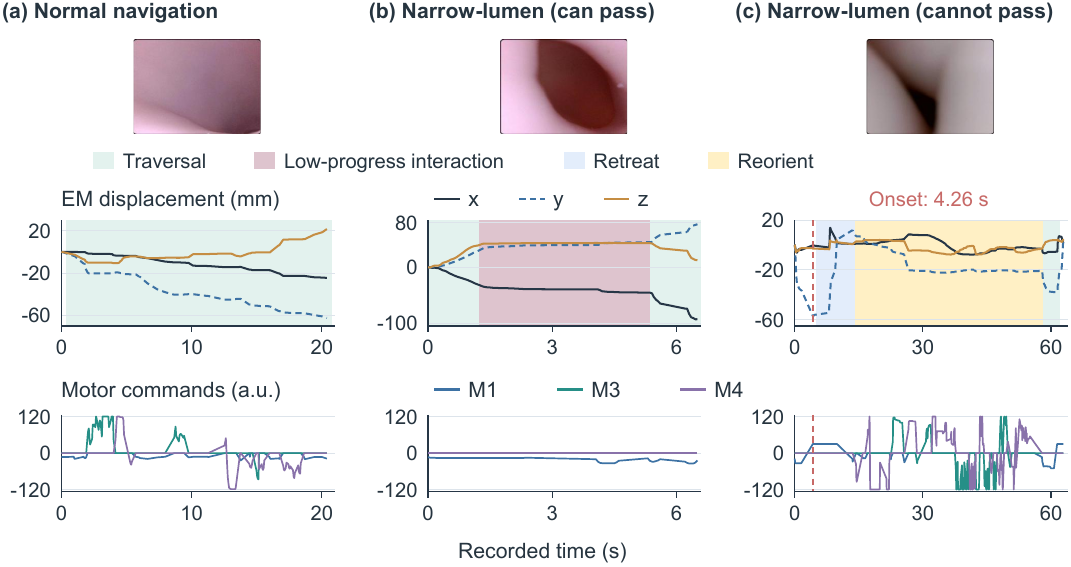}
\caption{Endoscopic views and synchronized recordings of (a) Normal navigation, (b) Narrow-lumen (can pass), and (c) Narrow-lumen (cannot pass), as defined by opening size relative to endoscope diameter. Shading reproduces the original qualitative phase annotations: traversal, low-progress interaction, retreat, and reorientation. EM traces show displacement from the initial position. Commands show M1, M3, and M4; M2 equals M1 in the saved logs. The red dashed line marks the logged first retreat onset at 4.256~s. Horizontal axes use recorded timestamps. Complete traces describe the recording context; predictive labels use controller transitions and pre-onset segments.}
\label{fig:examples}
\end{figure*}

\subsection{Matched Models and Evaluation}
We train six models on the original split with seeds 17, 42, and 2701: video (V), commands+EM (A+E), commands+video (A+V), EM+video (E+V), raw A+E+V, and MPRF. Controls share the temporal encoder, head, and fitting recipe. Stage A uses AdamW for 12 epochs at $10^{-3}$; Stage B uses up to 18 epochs at $3\times10^{-4}$, patience five on validation AUPRC, batch size 128, and gradient clipping at one. Both stages use weight decay $10^{-4}$. Training windows use stride two; evaluation uses stride one.

AUPRC is the primary metric for the sparse positive class. We also report AUROC, F1 and MCC at threshold 0.5, and Brier score. Seed SD captures fitting variability. Paired 95\% percentile intervals use 2,000 whole-episode resamples, stratified by recording condition; each draw shares episode multiplicities across models and averages over the three fitted seeds.

For alarms, validation considers thresholds 0.05--0.95 and persistence of one to five rows. All MPRF runs use the saved rule with threshold 0.95 and persistence five. We report the achieved evaluation burden. Event recall and false alarms per nominal minute (FAPM) use both 10- and 20-row matching windows applied to the same saved scores and alarms.

\section{Results}
\subsection{Why Intent Requires Response and Scene Context}
Figure~\ref{fig:examples} grounds the sensing problem in actual recordings. Normal navigation produces sustained displacement with intermittent commands. The passable narrow-lumen example includes a plateau during continued insertion, followed by renewed motion. The impassable example proceeds to retreat and steering. Thus reduced progress has a temporal context: its behavioral meaning depends on what follows and on the operator's ongoing intent.

\begin{figure*}[p]
\centering\includegraphics[width=\textwidth]{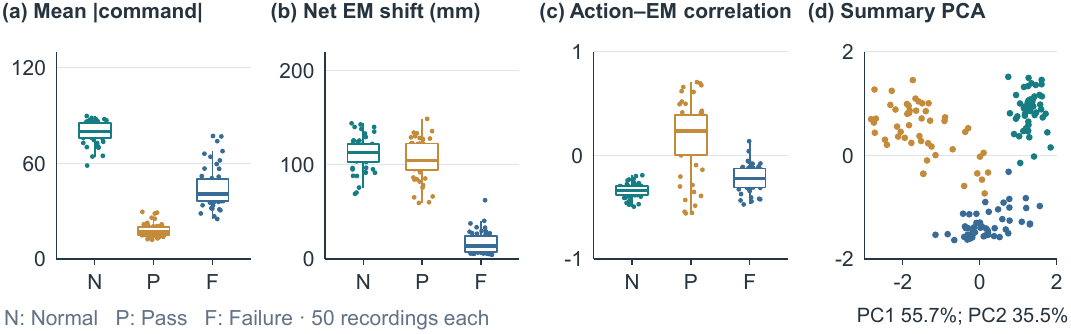}
\caption{Full-recording characterization, with one point per episode and boxplots showing median, interquartile range, and 1.5-IQR whiskers. (a) Mean rowwise maximum absolute command. (b) Net tip translation, $\|q_T-q_0\|_2$, using EM position $q$. (c) Pearson correlation between current command magnitude and incremental EM distance, over rows $1,\ldots,T$. (d) PCA of these three standardized summaries. N/P/F denote Normal/Pass/Failure. These descriptive quantities summarize complete recordings, including recovery.}
\label{fig:characteristics}
\end{figure*}

The full-recording analysis reinforces this ambiguity (Fig.~\ref{fig:characteristics}). Median command magnitudes are 79.99, 16.79, and 40.68 for Normal, Pass, and Failure, whereas median net EM translations are 113.09, 104.33, and 13.46~mm. Normal and Pass reach comparable final displacement under very different command magnitudes. Failure has an intermediate command level yet a small final shift, consistent with the return motion included in those recordings. The median command--increment correlation changes from 0.237 in Pass to $-0.223$ in Failure. The first two summary PCs explain 91.3\% of variance. Together, these descriptive results motivate contextual sensing; the following censored-window comparisons test anticipation directly.

\subsection{Multimodal Complementarity and the Residual Contribution}
\begin{figure*}[p]
\centering\includegraphics[width=\textwidth]{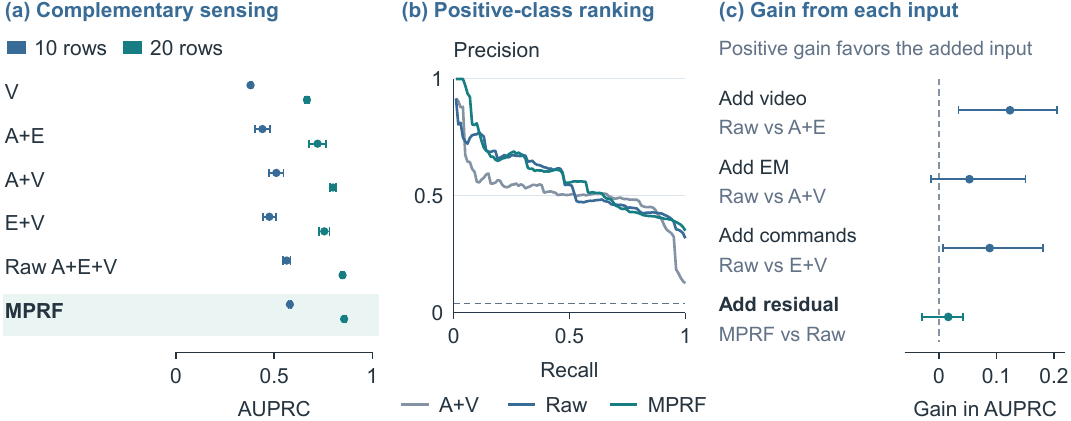}
\caption{Evidence for complementary multimodal sensing. (a) AUPRC at the trained 10-row target and 20-row timing tolerance (double the 10-row window); higher is better, and whiskers show seed SD. (b) Mean precision at a common recall grid across three seeds; the dashed line marks 10-row positive prevalence. (c) Mean AUPRC gains from adding video, EM, or commands to the respective two-modality control, followed by adding the residual to raw A+E+V fusion. Positive values favor the added input; bars show 95\% episode-bootstrap intervals. All three modality-removal controls have lower mean AUPRC, while MPRF versus Raw has the smallest mean gap, supporting the value of retaining all three modalities. The EM and residual intervals include zero.}
\label{fig:ranking}
\end{figure*}

Raw A+E+V achieves 0.564 AUPRC, compared with 0.580 for MPRF (Table~\ref{tab:ranking}). Removing video, EM, or commands lowers the mean to 0.440, 0.511, or 0.476. These gaps (0.124, 0.053, 0.088) each exceed the MPRF--Raw gap of 0.016 (Fig.~\ref{fig:ranking}c), identifying complete multimodal input as the principal source of the ranking advantage. MPRF exceeds A+V, the strongest reduced-modality control, by 0.069 [0.016, 0.146].

\begin{table*}[p]\centering
\caption{Matched 10-row anticipation results on 5,392 evaluation endpoints. AUPRC is mean $\pm$ SD across three seeds; other values are seed means. Bold identifies the best unrounded mean in each column. F1 and MCC use sample threshold 0.5. A: commands; E: EM; V: video.}
\label{tab:ranking}
\renewcommand{\arraystretch}{1.1}
\setlength{\tabcolsep}{11pt}
\begin{tabular}{lccccc}\toprule
Model & AUPRC $\uparrow$ & AUROC $\uparrow$ & F1 $\uparrow$ & MCC $\uparrow$ & Brier $\downarrow$\\\midrule
V & $0.380\pm0.006$ & $0.963$ & $0.482$ & $0.513$ & $0.061$\\
A+E & $0.440\pm0.040$ & $0.964$ & $0.465$ & $0.496$ & $0.065$\\
A+V & $0.511\pm0.037$ & $0.968$ & $0.545$ & $0.570$ & $0.050$\\
E+V & $0.476\pm0.034$ & $0.958$ & $0.539$ & $0.547$ & $0.047$\\
Raw A+E+V & $0.564\pm0.020$ & $0.977$ & $\mathbf{0.571}$ & $\mathbf{0.596}$ & $0.047$\\
MPRF (ours) & $\mathbf{0.580\pm0.007}$ & $\mathbf{0.977}$ & $0.551$ & $0.570$ & $\mathbf{0.046}$\\
\bottomrule\end{tabular}\end{table*}

\begin{figure*}[t]
\centering\includegraphics[width=\textwidth]{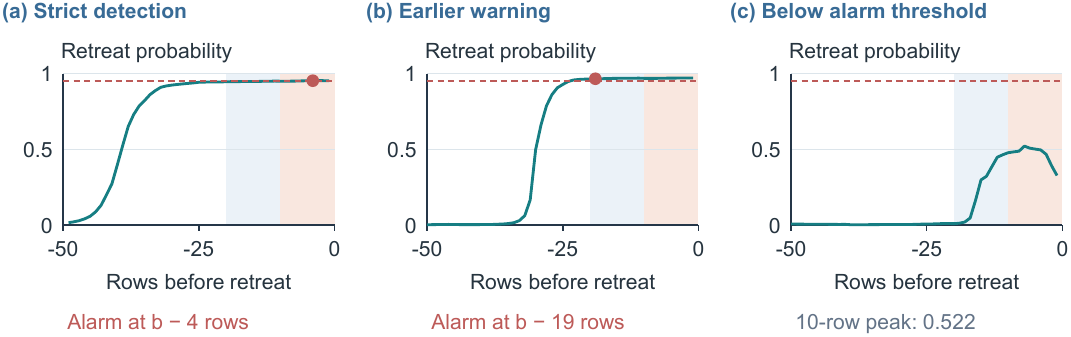}
\caption{Seed-17 MPRF traces under threshold 0.95 (dashed line), persistence five, and refractory interval 20 rows. Red points mark emitted alarms. Orange shading denotes the trained final 10 rows; blue shading adds rows 11--20. (a) A warning four rows before retreat counts under both definitions. (b) A warning 19 rows before retreat enters the expanded window. (c) The final-window peak remains 0.522. Cases are failure episode 13, attempts 0/1, and episode 8, attempt 1, respectively.}
\label{fig:alarms}
\end{figure*}

\newpage
Paired intervals are positive for Raw versus A+E (0.124 [0.034, 0.206]) and E+V (0.088 [0.007, 0.182]). Intervals for Raw versus A+V (0.053 [$-0.014$, 0.150]) and MPRF versus Raw (0.016 [$-0.029$, 0.041]) include zero. These results support multimodal complementarity with a smaller, uncertain residual contribution. Raw fusion leads F1 and MCC at threshold 0.5; MPRF leads mean AUPRC, AUROC, and Brier score, separating ranking and calibration from one fixed classification threshold.

\subsection{Alarm Timing Determines Event Attribution}
The frozen MPRF gate reaches 15.9\% recall at 2.78 FAPM in the trained 10-row window (Table~\ref{tab:alarms}). With 20-row matching, the same alarms reach 54.0\% recall at 0.97 FAPM. Raw fusion changes from 12.7\% at 2.86 FAPM to 44.4\% at 1.37 FAPM. MPRF's 20-row AUPRC is 0.857, compared with 0.849 for raw fusion. Expanded matching broadens the behavioral target and raises positive prevalence to 7.75\%; the predictor and gate stay fixed. The comparison therefore measures sensitivity to the retreat timestamp.

\begin{table}[t]\centering
\caption{Frozen alarms under two matching windows. Recall is in percent; FAPM uses nominal 20-Hz exposure. Bold marks the best mean across all six models.}
\label{tab:alarms}
\small\renewcommand{\arraystretch}{1.08}
\setlength{\tabcolsep}{3.5pt}
\begin{tabular}{lrrrr}\toprule
 & \multicolumn{2}{c}{10 rows (0.5 s)} & \multicolumn{2}{c}{20 rows (1.0 s)}\\
\cmidrule(lr){2-3}\cmidrule(lr){4-5}
Model & Recall$\uparrow$ & FAPM$\downarrow$ & Recall$\uparrow$ & FAPM$\downarrow$\\\midrule
V & 11.1 & 2.39 & 30.2 & 1.53\\
A+E & 6.3 & \textbf{1.62} & 19.0 & 1.05\\
A+V & \textbf{25.4} & 2.32 & 44.4 & 1.45\\
E+V & 7.9 & 3.16 & 47.6 & 1.29\\
Raw A+E+V & 12.7 & 2.86 & 44.4 & 1.37\\
MPRF (ours) & 15.9 & 2.78 & \textbf{54.0} & \textbf{0.97}\\
\bottomrule\end{tabular}\end{table}

Across three seeds, 10 alarms lie within 10 rows of retreat, 24 lie 11--20 rows before it, and 12 occur earlier. These are repeated evaluations of the same physical attempts. All 36 strict false alarms precede a later retreat; their timing explains the change in event recall.

Figure~\ref{fig:alarms} contrasts an alarm four rows before onset, an earlier alarm at 19 rows followed by refractory suppression, and a final-window peak of 0.522. Advisory timing depends on both score evolution and the gate.

\section{Discussion and Conclusion}
MPRF connects the expected--actual motion discrepancy to operator retreat through causal multimodal histories. Complementary sensing provides the strongest evidence; the residual adds a smaller, uncertain ranking gain.

Retrospective evaluation on a fixed single-operator phantom split measures agreement with recorded retreat timing. Operator habits, risk tolerance, and reorientation strategies differ, making this timestamp a behavioral proxy. Timing agreement is an initial learning endpoint; the robotics objective is multimodal inference of human-like navigation intent that improves task success. Expanded matching windows illustrate this distinction by changing scores for identical warnings.

Planned real-robot studies will use success at prespecified reachable navigation goals as the primary endpoint across operators and lumen conditions. Added distal-tip force sensing will compare MPRF-guided retreat with baseline navigation. Peak force and the duration and fraction of time above a prespecified high-force threshold will test whether retreat reduces elevated-load exposure. These task and safety endpoints will quantify the practical value of learned intent judgment.

\section*{Acknowledgement}
ChatGPT (OpenAI) assisted with language revision and icon generation. The authors reviewed the resulting material and remain responsible for its scientific content.
\clearpage
\bibliographystyle{IEEEtran}
\bibliography{references}
\end{document}